\pdfoutput=1

\documentclass[11pt]{article}

\usepackage{acl}

\usepackage{times}
\usepackage{latexsym}
\usepackage{xcolor}
\usepackage[T1]{fontenc}

\usepackage[utf8]{inputenc}

\usepackage{microtype}

\usepackage{inconsolata}

\usepackage{graphicx}
\usepackage{hyperref}

\usepackage{tikz,graphics,color,fullpage,float,epsf,caption,subcaption,float,amsmath,multirow,booktabs,bbding,pifont,listings,changepage,tabularx}
\usepackage[normalem]{ulem}

\definecolor{green}{rgb}{0.345, 0.627, 0.333}
\newcommand{\green}[1]{\textcolor{green}{#1}}

\title{InstructCoder: Instruction Tuning Large Language Models for Code Editing}

\author{
Kaixin Li$^{1}$\thanks{~Equal contribution. Ordering is determined by dice rolling.}  \quad Qisheng Hu$^{1}$\footnotemark[1] \quad Xu Zhao $^{1}$ \quad Hui Chen $^{2}$ \quad Yuxi Xie $^{1}$ \quad Tiedong Liu $^{1}$  \\ 
\textbf{Qizhe Xie}$^{1}$\thanks{~\hspace{1pt}Equal advising. Ordering is determined by dice rolling.}
\quad \textbf{Junxian He}$^{3}$\footnotemark[2]  \\
  $^1$National University of Singapore \quad $^2$Singapore University of Technology and Design \\ \quad $^3$Shanghai Jiao Tong University \\
 \texttt{\{likaixin,qishenghu,xu.zhao,xieyuxi,tiedong.liu\}@u.nus.edu},  \\ \texttt{hui\_chen@mymail.sutd.edu.sg}, \\ \texttt{junxianh@sjtu.edu.cn} \\
  }

\begin{document}
\maketitle
\begin{abstract}
Code editing encompasses a variety of pragmatic tasks that developers deal with daily. Despite its relevance and practical usefulness, automatic code editing remains an underexplored area in the evolution of deep learning models, partly due to data scarcity. In this work, we explore the use of Large Language Models (LLMs) to edit code based on user instructions. Evaluated on a novel human-written execution-based benchmark dubbed \textbf{EditEval}, we found current models often struggle to fulfill the instructions. In light of this, we contribute \textbf{InstructCoder}, the first instruction-tuning dataset designed to adapt LLMs for general-purpose code editing, containing high-diversity code-editing tasks such as comment insertion, code optimization, and code refactoring. It consists of over 114,000 instruction-input-output triplets and covers multiple distinct code editing scenarios. The collection process starts with filtered commit data sourced from GitHub Python repositories as seeds. Subsequently, the dataset is systematically expanded through an iterative process, where both seed and generated tasks are used to prompt ChatGPT for more data. Our findings reveal that open-source LLMs fine-tuned on InstructCoder can significantly enhance the accuracy of code edits, exhibiting superior code-editing performance matching advanced proprietary LLMs. 


The dataset and the source code are available at \url{https://github.com/qishenghu/CodeInstruct}. 
\end{abstract}

\section{Introduction}



     
\begin{figure*}[t!]
     \centering
         \includegraphics[width=1.\textwidth]{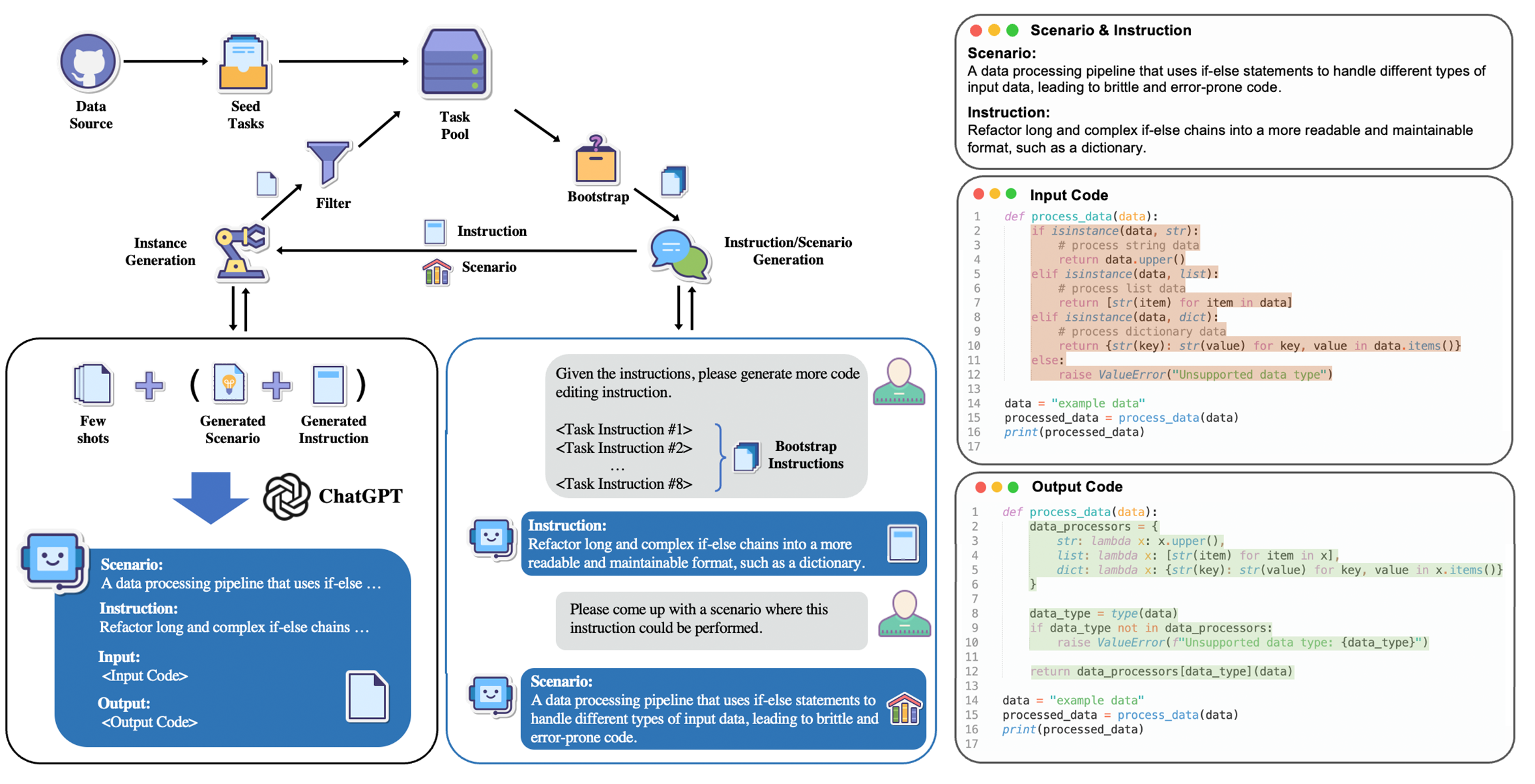}
        \caption{
        Data collection pipeline of InstructCoder (left) and a qualitative example from the dataset (right, best viewed with zoom). Initial seed tasks are selected from GitHub commits, and inspire ChatGPT to generate new instructions. Plausible scenarios where the filtered instructions may be used are then generated. Finally, corresponding code input and output are obtained conditioned on both the instruction and scenario. High-quality samples are manually selected and recurrently added to the task pool for further generation.}
         \label{fig:flow_chart}
\label{fig:main}

\end{figure*}

Developers typically engage in a cyclic routine of writing and revising code. As a crucial element, code editing takes up a great portion of this process, encapsulating diverse sub-tasks such as code optimization, refactoring, and bug fixing, each posing distinct challenges. Automated code editing tools could substantially boost developer productivity by alleviating the burden of monotonous tasks. However, it remains an under-explored area, partly due to the lack of relevant data, hampering substantial progress by deep learning models.


Inspired by the recent advancements in LLMs~\cite{brown2020GPT3,chowdhery2022palm,ouyang2022InstructGPT,2022ChatGPT,touvron2023llama,2023GPT4} and Code LLMs~\citep{nijkamp2023codegen2, chen2021codex, li2023starcoder}, we explore the proficiency of LLMs in code editing tasks based on user instructions, for instance, ``add a docstring to the function for clarity'', ``remove redundant code'', or ``refactor it into reusable functions''. These tasks are distinctly different from code completion, which involves generating code to complete given code snippets or comments. Code editing requires the model to not only understand the existing code but also execute modifications that are in line with the given instructions, while seamlessly integrating with the context. For example, removing redundant code or refactoring a function should not affect the return value.

To systematically evaluate LLMs for code editing, we created a novel benchmark named \mbox{\textbf{EditEval}}. It contains various types of code edits adapted from Github commits and existing datasets. Intriguingly, we found that open-source models yield unsatisfactory results, and even the most advanced proprietary LLMs struggle to solve these tasks.

In addressing this challenge, we present InstructCoder, a diverse dataset for instruction finetuning, particularly designed to improve the code editing abilities of LLMs. 
Specifically, we first collect and manually scrutinize commit data from public repositories on GitHub as the seed code editing tasks. Then, we utilize the seed data to prompt ChatGPT~\citep{2022ChatGPT} to generate new instructions and input-output pairs respectively.
This process resembles the Self-Instruct~\citep{wang2022self-instruct} and Alpaca~\citep{2023alpaca} frameworks. 
By innovatively forcing scenarios to guide the generation process, our approach ensures that the tasks in InstructCoder are diverse and relevant to real-world programming situations, 
resulting in a robust dataset for instruction finetuning in the code editing domain. After proper deduplication and postprocessing, we retain over 114,000 samples in the dataset. 

Our empirical studies reveal that LLMs display notable gains in code editing abilities after fine-tuning with InstructCoder. Code LLaMA achieves the best results through fine-tuning, attaining an accuracy of 57.22\%, closely matching ChatGPT.
Further studies also signify that while the pre-training of the models is fundamental, the code editing performance is highly influenced by the quality and volume of the instruction-tuning data.

In summary, the contributions of this work are (1) \textbf{InstructCoder}, the first instruction-tuning dataset featuring a wide range of diverse code editing tasks, and demonstrate the effectiveness of instruction-finetuning with InstructCoder; (2) \textbf{EditEval}, a novel human-written execution-based benchmark for the rigorous evaluation of general-purpose code editing; (3) We find that open-source models instruction-tuned with InstructCoder can demonstrate strong code editing performance matching ChatGPT.

\begin{figure*}[t!]
     \centering
         \includegraphics[width=\textwidth]{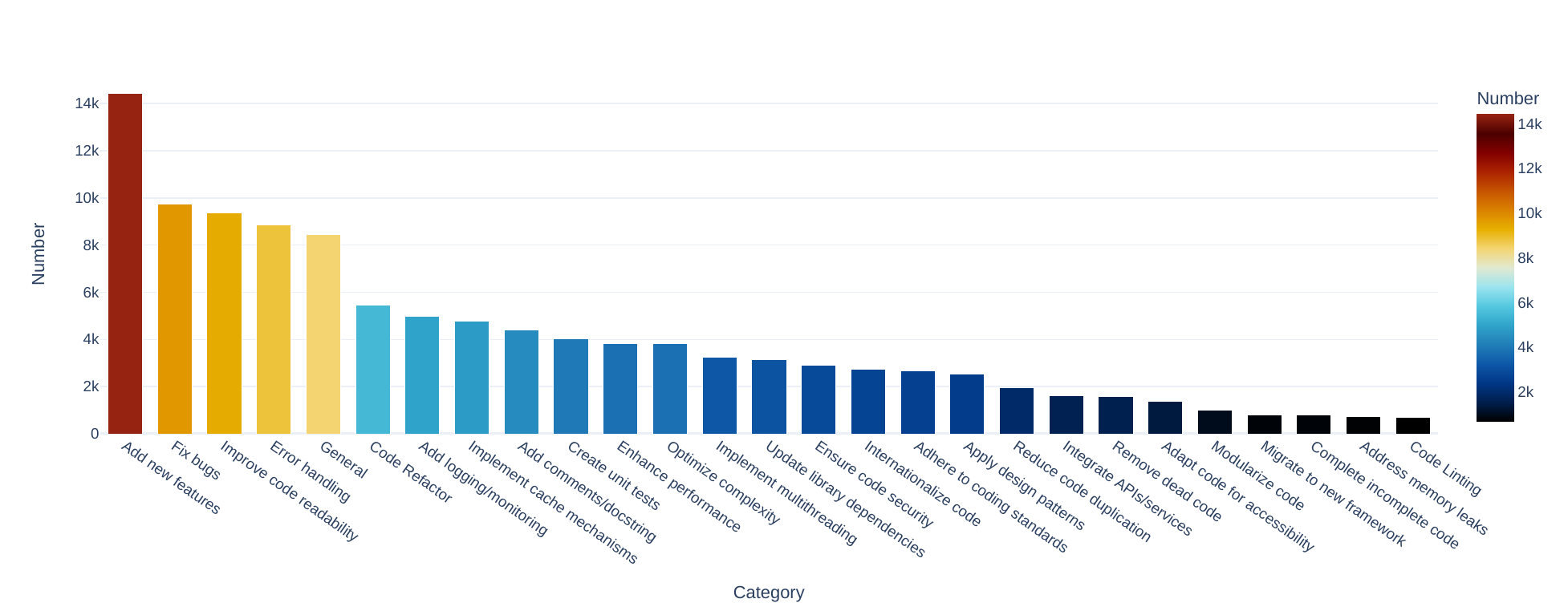}
        \caption{Distribution of code edit intent categories.}
         \label{fig:intent_distribution}

\end{figure*}

\begin{figure*}[t!]


     \centering
     \begin{subfigure}[b]{0.48\textwidth}
         \centering
         \includegraphics[width=0.85\textwidth]{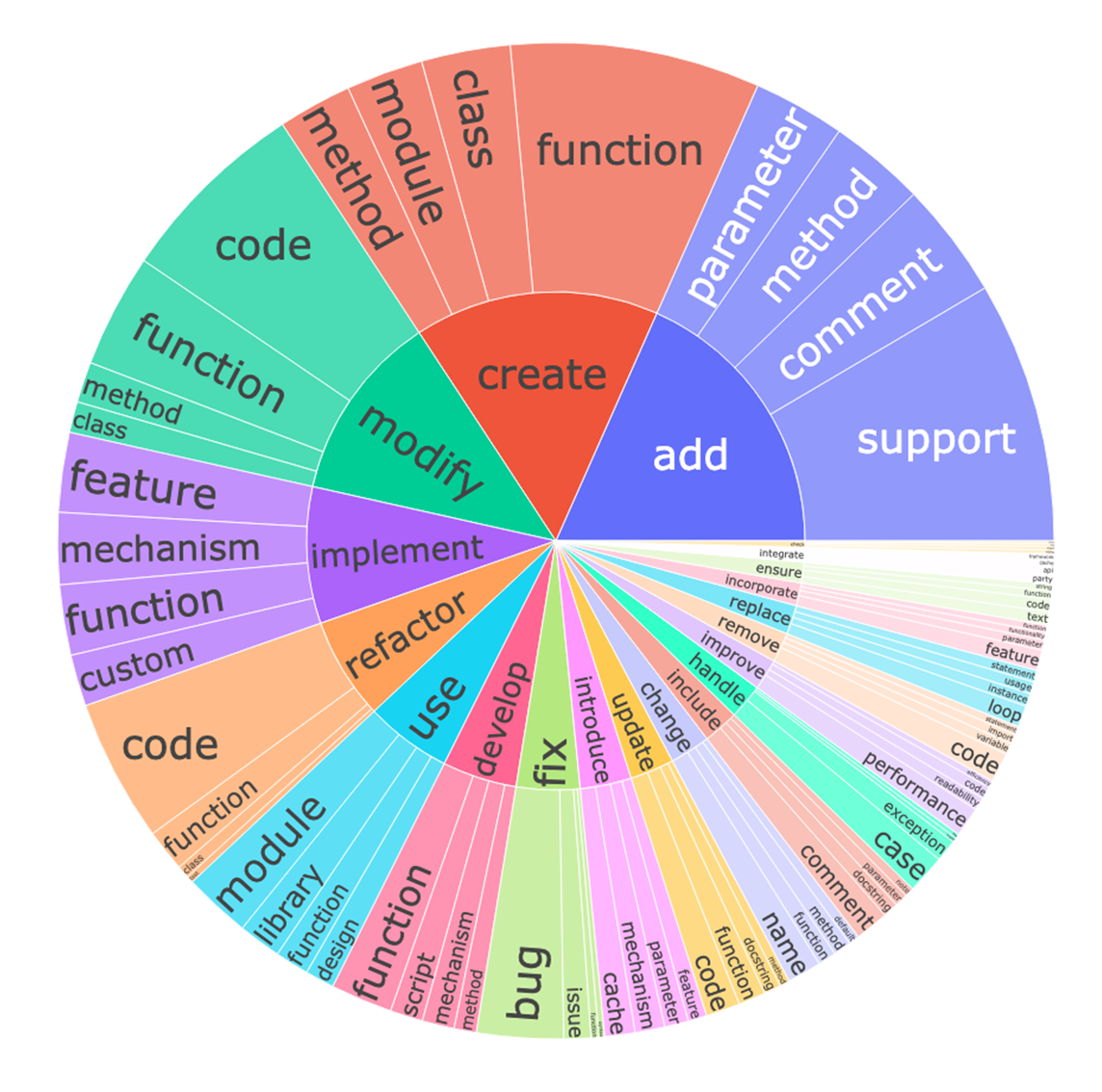}
         \caption{The top 20 most common root verbs with each top 4 noun objects in the instructions. Instructions with other infrequent root verbs take up 25\%.}
         \label{fig:verb_noun}
     \end{subfigure}
     \hfill
    \begin{subfigure}[b]{0.48\textwidth}
         \centering
         \includegraphics[width=0.85\textwidth]{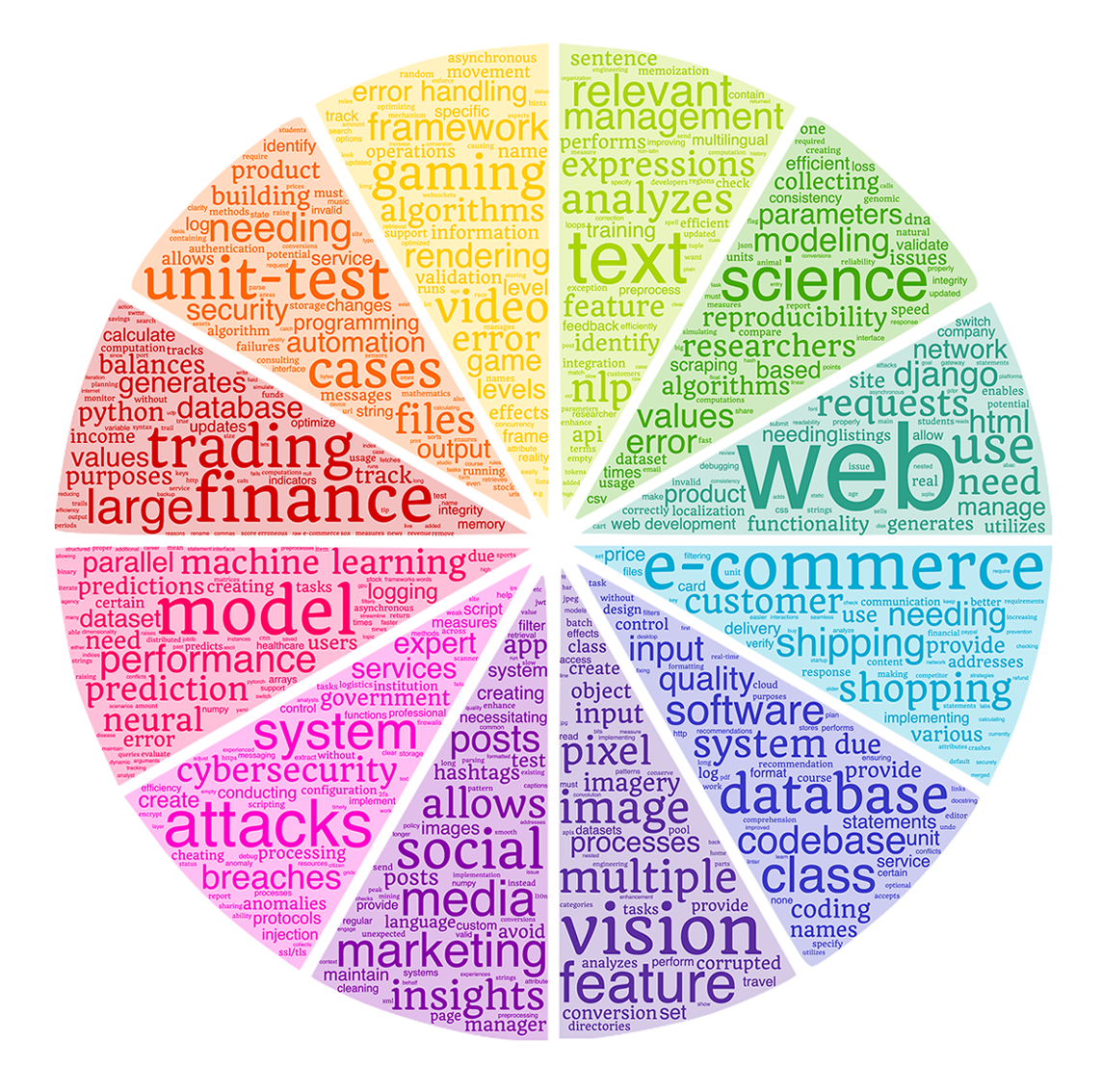}
         \caption{Wordcloud of scenario domains. Each sector with a different color corresponds to a different scenario domain. Each domain is a cluster of similar scenarios.}
         \label{fig:scenario_word_cloud}
     \end{subfigure}

        \label{fig:three graphs}
\caption{Visualizations of InstructCoder data. Best viewed in zoom.}
\label{fig:data_distribution}

    
\end{figure*}

\section{Related Work}
\subsection{Instruction Finetuning Datasets}
Previous studies have concluded that instruction finetuning LLMs on a diverse collection of instructional tasks can further improve the ability of LLMs to generalize well on unseen tasks~\citep{ouyang2022InstructGPT,mishra-etal-2022-cross,wei2021flan,2022flan-t5,wang2023multi-task-learning}. To support these tasks, datasets consisting of a large number of code snippets with corresponding annotations are necessary. 
These instruction can be reformulated from existing datasets~\citep{aribandi2021exmix,wei2021flan,mishra-etal-2022-cross,longpre2023flan2023}, or human-written with crowd-sourcing efforts~\citep{ouyang2022InstructGPT,wang-etal-2022-super}.  Machine generation of instruction data has also been explored to reduce human labour~\citep{wang2022self-instruct,honovich2022unnatural-instruct,2023alpaca,2023instructionwild}. Despite the presence of elevated noise levels within the data, its effectiveness has been identified.

\subsection{Code Synthesis}
Code generation is an extensively studied area. Language models pre-trained on large collections of code have demonstrated strong abilities in a variety of programming tasks. Some general LLMs gain code generation abilities due to the mixture of code in the pre-training corpus (e.g. The Pile~\citep{gao2020pile}), such as GPT-3~\citep{brown2020GPT3}, ChatGPT, GPT-4~\citep{2023GPT4}, LLaMA~\citep{touvron2023llama}, BLOOM~\citep{scao2022bloom}, GPT-NeoX~\citep{black2022gpt-neox}, and Pythia~\citep{biderman2023pythia}. LLMs specifically trained on code and optimized for code generation are also studied, e.g. Codex~\citep{chen2021codex}, CodeGen~\citep{nijkamp2023codegen}, CodeGeeX~\citep{zheng2023codegeex} and StarCoder~\citep{li2023starcoder}. These models all adopt the decoder-only transformer architecture but differ in size and specific model design (e.g. positional embedding, norm layer placement) as well as the selection and preprocessing of the pre-training corpus. 

On the other hand, relatively little literature addresses the objective of code editing. Previous works focus on a subset of code editing tasks, such as code infilling~\citep{fried2022incoder} and debugging~\citep{just2014defects4j,tarlow2020learning-to-fix-build-errors,ding2020patching,jimenez2023swebench}. The PIE~\citep{madaan2023pie} dataset is a concurrent work most relevant to ours, which focuses on speeding up programs. Other works~\citep{yin2018learning-edit-repr,wei2023coeditor,chakraborty2020codit} can not accept natural language as edit intentions, rendering them less user-friendly.

Nevertheless, datasets particularly tailored for general-purpose code editing are absent. To fill this gap, we introduce InstructCoder, a novel dataset aimed at further advancing the capabilities of code editing with LLMs.

\section{EditEval: Evaluating Code Editing Models}

\begin{table}
\small
\centering

\begin{tabular}{lcccc}
\toprule
\textbf{Model} & \multicolumn{3}{c}{\textbf{Accuracy (\%)}} \\
\hline 
\noalign{\vspace{1mm}}
ChatGPT {\tiny (gpt-3.5-turbo-0613)} &  \multicolumn{3}{c}{57.73} \\
GPT-4 {\tiny (gpt-4-0613)} &  \multicolumn{3}{c}{68.56} \\
GPT-4 Turbo {\tiny (gpt-4-1106-preview)} &  \multicolumn{3}{c}{66.49} \\
\midrule
 & \textbf{7B} & \textbf{13B} & \textbf{33B} \\
Alpaca & 12.37 & 19.59 & 30.93 \\
LLaMA+CodeAlpaca & 18.56 & 18.56 & 35.56 \\
\bottomrule
\end{tabular}

\caption{Results of several instruction-tuned models evaluated on EditEval. }
\label{tab:editeval}
\vspace{-3mm}
\end{table}
As aforementioned, code editing is significantly different from code completion. Consequently, widely utilized datasets in the realm of code completion, such as MBPP~\citep{austin2021program} and HumanEval~\citep{chen2021evaluating}, fall short in evaluating code editing capabilities.
To rigorously evaluate the code editing capabilities, we curated a test set of 194 code editing tasks, derived from three key sources: GitHub commit data, MBPP, and HumanEval. 
We harness the input code from these sources and create plausible edit instructions. For GitHub sources, we manually create execution contexts so that the code is runnable. Each sample is accompanied by a canonical solution written by humans to ensure the instruction is viable.
The generated code edits are strictly assessed using automated test cases to evaluate the correctness of the edits. An edit is considered correct only if it passes all the test cases. This automated method provides a robust and objective evaluation framework, essential for benchmarking the model's performance in diverse code editing situations. Appendix \ref{appendix:example-editeval} showcases an example of the test set. 

We benchmarked several instruction-tuned models on EditEval, and the results are listed in Table \ref{tab:editeval}. Generally, the results reveal significant potential for improvement in code editing. Alpaca and CodeAlpaca exhibit accuracies below 20\% with 7B and 13B sizes, and it only gets better at 33B. At this size, CodeAlpaca beats Alpaca, achieving 35.56\% accuracy. Turning to the GPTs, the most advanced proprietary models up to this point, GPT-4 achieves the best performance at 68.56\%. Even ChatGPT struggles at this task, scoring only 57.73\%. Upon closer examination, we found the challenge of EditEval lies in the high demand for both instruction following and code understanding. The model has to have a grasp of the implicated context of the input code, and then accomplish the edit within its context.

\section{InstructCoder: Instruction-tuning Empowers Code Editing}
In this section, we introduce how we create InstructCoder to boost the code editing abilities of LLMs via instruction finetuning. We employed a method based on Self-Instruct~\citep{wang2022self-instruct}, which expands instruction finetuning data by bootstrapping off language model generation. The methodology of generating data with LLMs requires minimal human-labeled data as seed tasks while maintaining the quality and relevance of the tasks in the dataset. Through an iterative process of generating instructions and refining them with deduplication, we create a dataset of a wide range of code-editing tasks. Figure \ref{fig:main} illustrates the data collection pipeline of InstructCoder.

\subsection{Seed Data Collection}
\label{sec:seed_data_collection}
GitHub is a code hosting platform whose version control service naturally records code edits with commits, which can be converted to instructions. The repositories on GitHub provide diverse data with human-generated quality. However, the data is not suitable for direct utilization\footnote{Initial attempts to utilize real-world GitHub commit data for model fine-tuning yielded suboptimal results. Please refer to Appendix \ref{appendix:raw_data} for a detailed discussion.}. First, commit messages are mostly brief and resultant, missing detailed descriptions. Furthermore, they can be imprecise or even absent. Second, commits can be huge involving multiple files, which is beyond the scope of this work. In light of this, we direct our attention towards LLMs as a means to generate data, instead of the direct utilization of collected data.

Raw GitHub commit data were collated using BigQuery\footnote{\url{https://cloud.google.com/bigquery}}. To ensure high quality and address licensing issues, we focused on Python repositories on GitHub with over 100 stars and permissive licenses. Our selection criteria was restricted to commits modifying only one code block within a single Python file. These commits were identified by git-diff\footnote{\url{https://git-scm.com/docs/git-diff}}.

During the collection process, we came across many imprecise or emotionally charged commit messages. Codex~\citep{chen2021codex} was employed in such cases to clarify the changes made between versions and improve the commit messages, resulting in more precise and informative instructions. A total of 634 tasks were processed from the commit data through manual efforts and were used for the self-instruct process.

In addition to GitHub commit data, we also leverage high-quality generated samples as additional seed tasks. With manual inspection, a batch of 592 high-quality samples was compiled as additional seed tasks. This set of seed data covers a wide range of code-editing scenarios and enriches the basis on which InstructCoder is created, ensuring that the tasks are rooted in plausible real-world code-editing cases\footnote{Incorporating additional seeds also allows for modulating the distribution of generated data, facilitating customization for specific requirements.}.

\subsection{Instruction Bootstrapping}
Self-Instruct~\citep{wang2022self-instruct} is as an effective automated framework for instruction data generation. It works by iterative bootstrapping off LLM's generation, presenting a way to enrich the instructional dataset while maintaining task quality and relevance from a small set of human-evaluated seed tasks. We leveraged a similar approach to generate diverse code editing instructional data. In each iteration, seven seed task instructions and one ChatGPT-generated task instruction are sampled and combined as a few-shot context to prompt ChatGPT for more instructions. 
To generate more diverse and practically applicable instructions, we also generated tasks across multiple sub-domains by specifying the editing intent in the prompt provided. Relevant prompts used can be found in Table \ref{tab:prompts} in Appendix \ref{appendix:prompts}. 

\subsection{Scenario-conditional Generation}
\label{sec:scenario-cond-generation}
We originally found many generated samples share similar codebases despite different instructions and few-shot examples provided. Such similarity could largely diminish the dataset's value. 
Empirical analysis suggests the issue could be attributed to LLM generating general codebases for input/output snippets when insufficient context is provided.
As a countermeasure, we propose to introduce code editing scenarios for input/output code generation. We present some examples in Figure \ref{fig:scenario_example1},\ref{fig:scenario_example2},\ref{fig:scenario_example3} in Appendix \ref{appendix:example-scenario-conditional-generation}, where we generally observe that instances generated with scenario demonstrate higher quality in terms of richer context and code structure compared to those without. 

For each generated instruction, we first prompted ChatGPT to generate practical events as ``real-world'' scenarios where the editing instruction could be performed, and randomly select one for instance generation in the next step.
Subsequently, the LLM was instructed to generate samples that correspond with the instruction and scenario, ensuring the codebases and variable names are appropriate. The prompt used can be found in Table \ref{tab:prompts} in Appendix \ref{appendix:prompts}.

By incorporating scenario-conditional generation, the resulting samples exhibit increased variability regarding codebases and variable naming, thus augmenting the diversity of InstructCoder. 

\subsection{Postprocessing}
Following Self-Instruct~\citep{wang2022self-instruct}, deduplication was applied on the generated instructions to remove instructions that have a ROUGE-L~\citep{lin2004rouge} overlap score larger than 0.7 with the existing instructions. For the code, we employed MinHash with Locality Sensitive Hashing (LSH) indexing to remove instances with a Jaccard similarity greater than 0.75. 
Ultimately, InstructCoder comprises over 114,000 distinct code editing tasks. For experimental purposes, we designated 95\% of the tasks for training, while the remaining 5\% formed our validation set.




\section{Data Analysis}
We analyze InstructCoder in terms of 1) diversity, 2) complexity, and 3) correctness. We provide distribution and complexity analyses of the task instances. Finally, we demonstrate through human investigation that our data is highly reliable. 


\begin{figure}[t]
\centering
\includegraphics[width=0.5\textwidth]{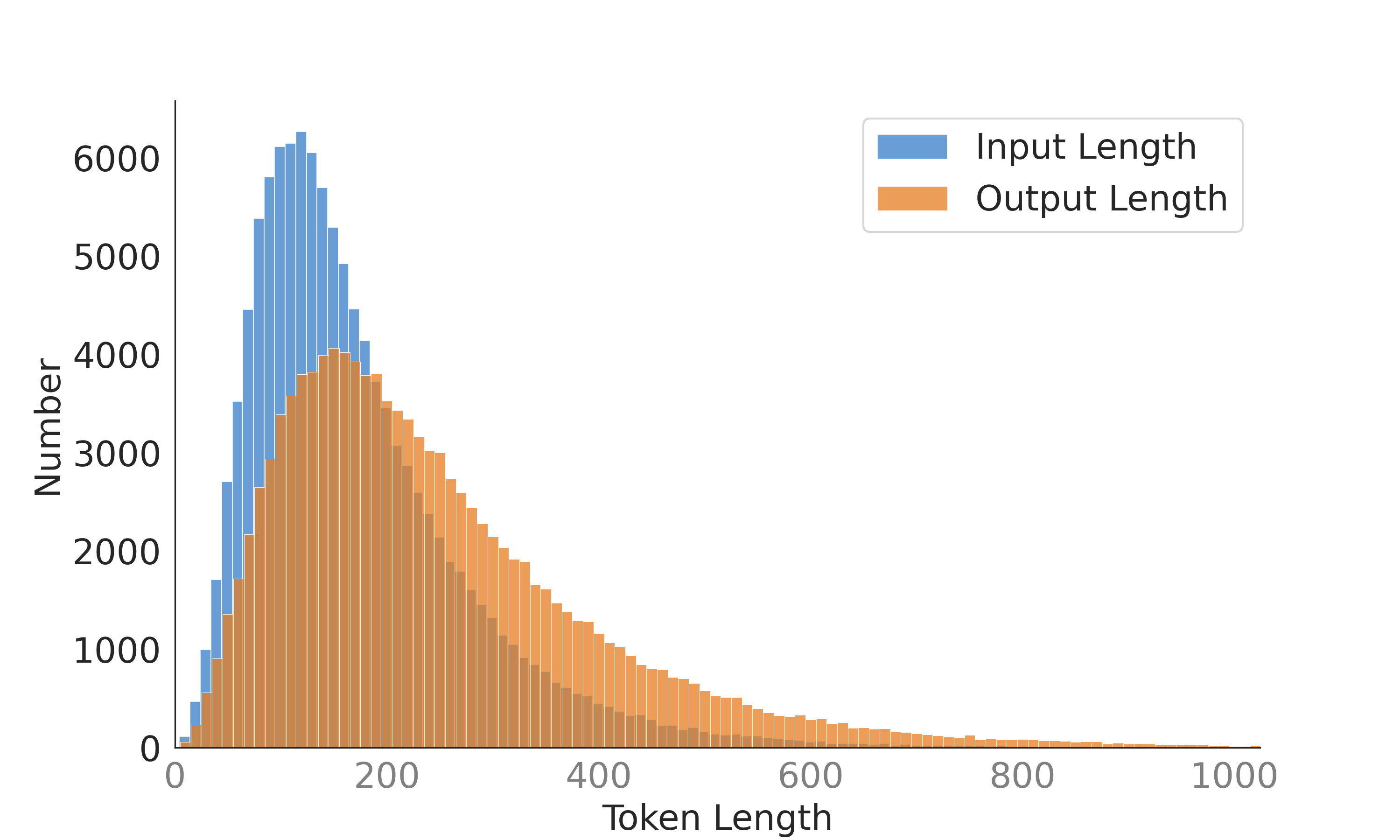}
\caption{Token length distribution of InstructCoder}
\label{fig:instance_length_distribution}
\end{figure}

\subsection{Statistic Overview}
InstructCoder comprises over 114k code editing instructions, each paired with an input/output instance. The token length distribution of input/output can be viewed in Figure \ref{fig:instance_length_distribution} and Table \ref{tab:statistic} in Appendix \ref{appendix:additional_stats}. 
Most of the data falls within a reasonable range in terms of length, while some extreme values reflect the breadth of our dataset.

\subsection{Instruction Diversity}

To explore the diversity of tasks in InstructCoder and their practical applicability, we present various instruction intents i.e. \emph{what} the code edits intend to accomplish, and instruction verbs, i.e. \emph{how} the code edit is accomplished.

\paragraph{Instruction Intents. } We asked ChatGPT to classify the types of code edits in our dataset and manually identified 27 empirical genres. Figure \ref{fig:intent_distribution} shows the distribution of the code edit intent categories in InstructCoder, which include adding functionalities, optimizing code, improving readability, etc. These objectives underscore the extensive range of InstructCoder.

\paragraph{Instruction Verbs. } The diversity of instruction verbs is also portrayed in Figure \ref{fig:verb_noun}. We demonstrate the top 20 root verbs and their top 4 direct nouns both ranked by frequency. While a great portion of the instructions can be roughly clustered as \emph{creation} (e.g. ``add'', ``implement'', ``creat'') and \emph{modification} (e.g. ``modify'', ``replace'', ``change''), InstructCoder presents a long-tail distribution with less common verbs other than the top-20 taking up 25.0\% percentage. This demonstrates that the dataset contains a wide spectrum of instructions.

\subsection{Scenario Diversity}
InstructCoder is designed to cover a wide range of scenarios. As discussed in Section \ref{sec:scenario-cond-generation}, each instruction was accompanied by different scenarios where the editing instruction could be performed to improve diversity. A word cloud is provided to show some of the scenario domains in our dataset, as illustrated in Figure \ref{fig:scenario_word_cloud}, with each sector referring to a different domain. The diversity of the dataset is emphasized by the presence of a wide range of domains such as image processing, web development, and cybersecurity.

\subsection{Complexity}
We reflect the complexity of a code edit task using the number of differing lines and their edit ratio in the input/output pair, which are defined as:
\begin{equation}
\mathit{n_{diff} = \vert I \cup O \: \backslash \: I \cap O \vert},
\end{equation}
\begin{equation}
\mathit{r_{diff} = \frac{n_{diff}}{\vert I \cup O \vert}},
\end{equation}
where $I$ and $O$ are sets of input/output code with single lines as elements.

We measure the differing lines of a code-editing task instance using the Python library \textit{difflib}.\footnote{\url{https://docs.python.org/3/library/difflib.html}} We found that the average number of differing lines in InstructCoder is 11.9 and the average edit ratio is 0.52. These values suggest a fairly acceptable level of complexity, indicating that the dataset is neither too easy nor too hard. InstructCoder strikes a balance in terms of complexity, making it well-suited for finetuning and evaluating LLMs in a wide range of code editing tasks.  Figure \ref{fig:edit_distribution} in Appendix \ref{appendix:additional_stats} illustrates the distribution of the number of differing lines.

\subsection{Correctness}
We further randomly sampled 200 instances and invite annotators to evaluate the instances based on two criteria: the validity of the instructions and the correctness of the outputs. The validity assessment focused on determining if the instructions exhibit clear and appropriate editing intents. The correctness evaluation examines if the input-output pairs reflect the changes specified by the instructions. 

The results in Table \ref{tab:quality_check} indicate that most instructions in the InstructCoder dataset are valid. A few instances exhibited noise and occasional failure to follow the instructions, but high correctness was found overall. Out of the 200 evaluated instances, 180 were successfully solved, showcasing the overall quality and reliability of InstructCoder.

\begin{table}
\small
\centering
\begin{tabular}{lll}
\toprule
\textbf{Question} & \textbf{Pass}\\
\hline
\begin{tabular}[t]{@{}l@{}}\small\verb|Determine if the instruction is valid.|\\\verb||\\\end{tabular} & 97$\%$ \\
\begin{tabular}[t]{@{}l@{}}\small\verb|Is the output an acceptable edited code|\\\small\verb|response to the instruction and input?|\end{tabular} & 90$\%$ \\
\bottomrule
\end{tabular}
\caption{Quality check questions and results on a randomly sampled subset with 200 data points.}
\label{tab:quality_check}
\vspace{-3mm}
\end{table}

\section{Experiments}

\subsection{Setup}
\label{sec:instruction_finetuning}
\paragraph{Training.}
We experiment with two families of open-source language models with various sizes: LLaMA (LLaMA, LLaMA-2 and Code LLaMA)~\citep{touvron2023llama,touvron2023llama2,roziere2023codellama} and BLOOM~\citep{scao2022bloom}.

LLaMA is a series of LLMs with parameters ranging from 7 to 65 billion. They have been pre-trained on a vast corpus, of which approximately 4.5\% comprises code. The LLaMA-2 series extends the family with more intensive pre-training. Additionally, Code LLaMAs are built on LLaMA-2 and specifically trained on 500B tokens of code to enhance its code understanding and generation capabilities. BLOOM is a multilingual LLM capable of generating human-like outputs in 46 languages and 13 programming languages.

A full finetuning updating all the parameters in an LLM can be computationally expensive. Instead, we adopt LoRA ~\citep{hu2021lora}, a parameter-efficient finetuning method that optimizes an approximated low-rank delta matrix of the fully-connected layers. 
In this way we could fine-tune a 33B model in a single A100-80GB GPU card. In our experiments, LoRA is applied to the query, key, value, and output transform weights of the Transformer architecture ~\citep{vaswani2017transformer}. All hyperparameters can be found in Table \ref{tab:hyperparameters} in Appendix \ref{appendix:hyperparameters}. 

\paragraph{Baselines.}
We select ChatGPT~\citep{2022ChatGPT}, GPT-4~\citep{2023GPT4} and GPT-4 Turbo as strong baselines. The aforementioned open-source models along with an instruction-tuned LLaMA model called Alpaca~\citep{2023alpaca} are included, and their zero-shot performance is reported.

Concurrent to our work, CodeAlpaca\footnote{\href{https://github.com/sahil280114/codealpaca}{https://github.com/sahil280114/codealpaca}} is a popular dataset generated with the pipeline of Alpaca, differing in that its seed data is replaced by hand-written easy instructions with short programs. We fine-tune LLaMA models with CodeAlpaca and Alpaca and compare the results.


\section{Results}


\subsection{Finetuning Efficacy with InstructCoder}

     

  

 


\begin{table}
\small
\centering
\scalebox{0.9}{
\begin{tabular}{lcccc}
\toprule
\multirow{2}{*}{\textbf{Model}} & \multirow{2}{*}{\textbf{Size}} & \multicolumn{2}{c}{\textbf{Accuracy (\%)}} & \multirow{2}{*}{\textbf{$\Delta$ Acc}} \\
& & \textbf{w/o ft} & \textbf{w/ ft} \\

\midrule
ChatGPT {\tiny (gpt-3.5-turbo-0613)} & - & \textbf{57.73} & - & - \\
\midrule
\multirow{2}{*}{BLOOM} & 3B & 0.52 & 15.46 & \green{+ 14.94}\\ 
 & 7B & 1.03 & 19.59 & \green{+ 18.56}\\
\noalign{\vspace{1mm}}
\multirow{3}{*}{LLaMA-1} & 7B & 2.57 & 26.80 & \green{+ 24.23}\\
 & 13B & 6.19 & 28.35 & \green{+ 22.16}\\
 & 33B & 6.19 & 41.75 & \green{\textbf{+ 35.56}}\\
 \noalign{\vspace{1mm}}
\multirow{2}{*}{LLaMA-2} & 7B & 4.12 & 27.32 & \green{+ 23.20}\\
 & 13B & 14.95 & 34.54 & \green{+ 19.59}\\
 \noalign{\vspace{1mm}}
\multirow{2}{*}{Code LLaMA} & 7B & 29.90 & 45.88 & \green{+ 15.98}\\
 & 13B & 28.86 & \textbf{57.22} & \green{+ 28.36}\\
\bottomrule
\end{tabular}
}
\caption{Models finetuned with InstructCoder significantly improve in code edit accuracy on EditEval, regardless of the model family or model size. }
\label{tab:main_results}
\vspace{-3mm}
\end{table}

\label{sec:results}

In this section, we demonstrate the value of our InstructCoder dataset. Table \ref{tab:main_results} presents a detailed comparison of EditEval performance across models fine-tuned with InstructCoder and baseline models.
While very low accuracies are observed in open-source plain models, finetuning with InstructCoder significantly boost the accuracy, highlighting the effectiveness of efficient instruction finetuning with machine-generated code edit pairs.

Code LLaMA 13B matches ChatGPT's performance and surpasses other open-source models with a 57.22\% accuracy rate. The more substantial LLaMA-33B model shows a notable 35.56\% improvement, yet it falls behind Code LLaMA-7B, which benefits from extensive pre-training on code. For qualitative results, see Appendix \ref{appendix:qualitative_examples}.

As expected, the pre-training foundation of LLM significantly influences code-editing efficacy. LLaMA demonstrated higher accuracies than BLOOM models of similar sizes. Among LLaMAs, those pre-trained on more tokens (LLaMA-2 series) outperformed earlier versions. Furthermore, Code LLaMAs exceed LLaMA-2 models as a result of their extensive pre-training specifically on coding data. Despite the varying capabilities of the foundational models, our dataset consistently enhances performance.

\begin{figure}[t!]
\centering
\includegraphics[width=0.45\textwidth]{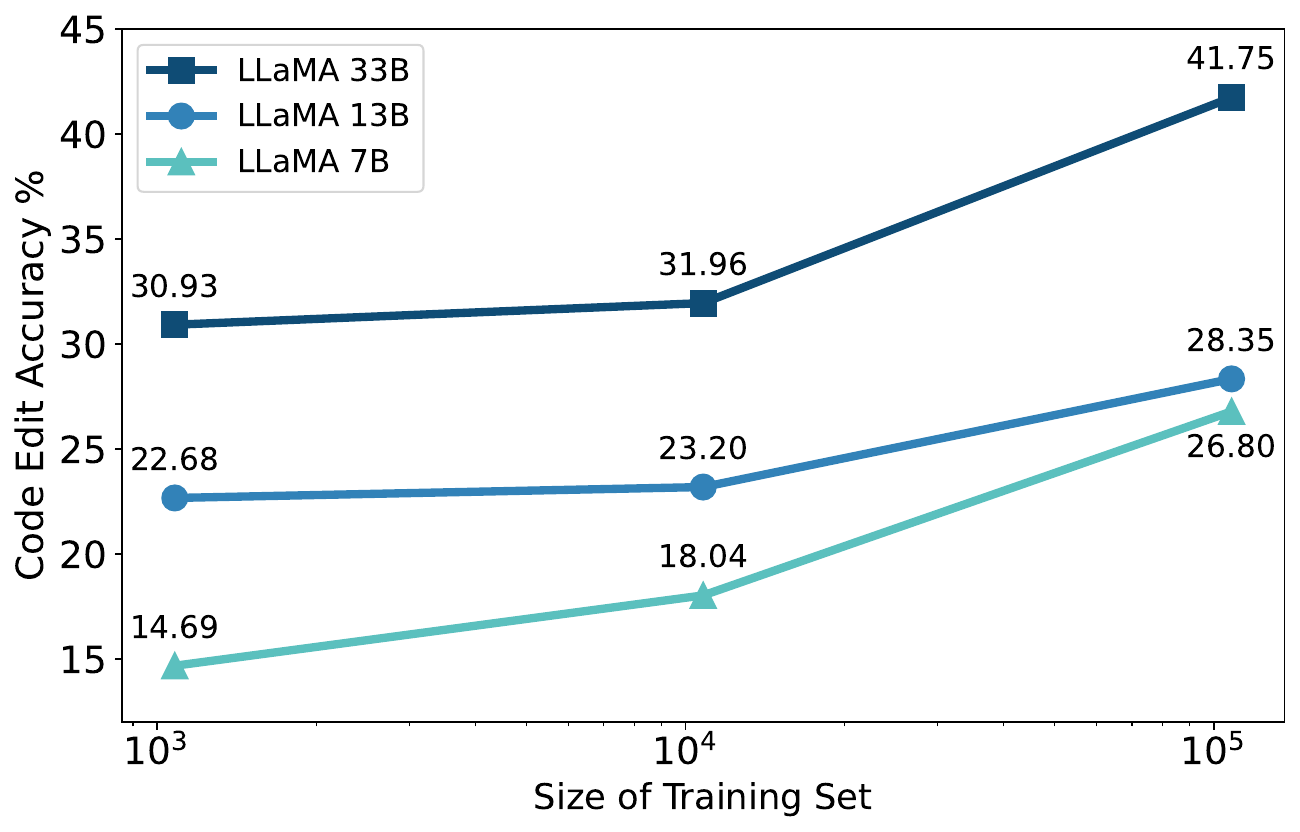}
\caption{Data scaling performance of InstructCoder on LLaMA evaluated on EditEval, using 1\%, 10\% and 100\% training data.}
\vspace{-5mm}
\label{fig:data_scaling}
\end{figure}

\subsection{Dataset Scaling}
InstructCoder has a scale considerably smaller than what LLMs are typically pre-trained on. To ascertain the sufficiency of this scale, we conducted an experiment wherein we fine-tuned the LLaMA models using varying proportions (1\%, 10\%, and 100\%) of the dataset. The smaller subsets are guaranteed to be encompassed within the larger subsets. The results are shown in Figure \ref{fig:data_scaling}.
The identified trend demonstrates a positive correlation between the model's accuracy and the scale of the training set.

Fine-tuned with merely 1\% of the data, the models experience a limited number of parameter updates but quickly adapt to the tasks, surpassing their respective zero-shot accuracy scores by significant margins. This underscores the significance of instruction tuning. As the volume of training data increases, we observe consistent improvements in model accuracy, approximately growing linearly with respect to the logarithmic scale of the number of samples. 
Crucially, our experiment empirically suggests that larger models are more effective with a constrained training compute budget.

\subsection{Edit Ratio}

\begin{figure}[t]
\centering
\includegraphics[width=0.45\textwidth]{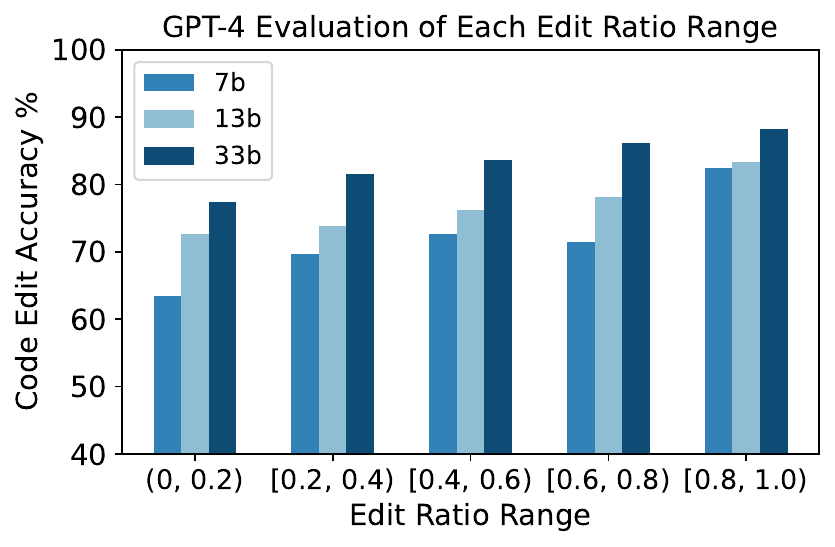}
\caption{GPT-4 evaluation results at different edit ratios on 2000 validation samples.}
\label{fig:acc_edit_ratio}
\vspace{-2mm}
\end{figure}

Figure \ref{fig:acc_edit_ratio} depicts the accuracy of fine-tuned LLaMA models as evaluated by GPT-4 across five edit ratio levels, using 2000 random samples from the validation set. This evaluation, justified in Appendix \ref{appendix:discussion_gpt-4-eval}, involves prompting GPT-4 for a quick and general assessment of code edits, offering an alternative perspective to code edit evaluation. In this assessment, larger models consistently outperform their smaller counterparts. Notably, accuracy decreases with lower edit ratios, potentially due to the models adopting the shortcut of copying inputs to minimize loss in scenarios requiring fewer edits. This trend, however, is less pronounced in larger models, which show a greater ability to discern subtle differences in cases of low edit ratios.

\section{Conclusion}
We introduce InstructCoder, the first instruction-tuning dataset for general-purpose code-editing tasks. It comprises generations of LLMs, where real GitHub commits serve as seed tasks to guide the generation process. A scenario-conditional approach is introduced to ensure both diversity and high quality of the data. Our experiments on the novel EditEval benchmark show that open-source models can gain huge improvements and even yield performance matching proprietary models through computationally lightweight parameter-efficient fine-tuning with InstructCoder. We also reveal that the LLM base model and the scale of fine-tuning data are both profound factors of code-editing ability. We hope the dataset can benefit and inspire more research in this area towards building more powerful coding models.

\section*{Limitations}
Our approach did not encompass code changes involving multi-file contexts, which might be useful in development. We hope to explore these aspects further and incorporate additional programming languages in our future research. 



\bibliography{refs}


\newpage
\onecolumn
\appendix

\section{An Example Test of EditEval}
\label{appendix:example-editeval}

An example of EditEval the test set is showcased below. 
To accomplish the task, the model must not only adhere to the user's instructions but also comprehend the input code in the context provided.

\begin{figure*}[h!]
\centering
\includegraphics[width=\textwidth]{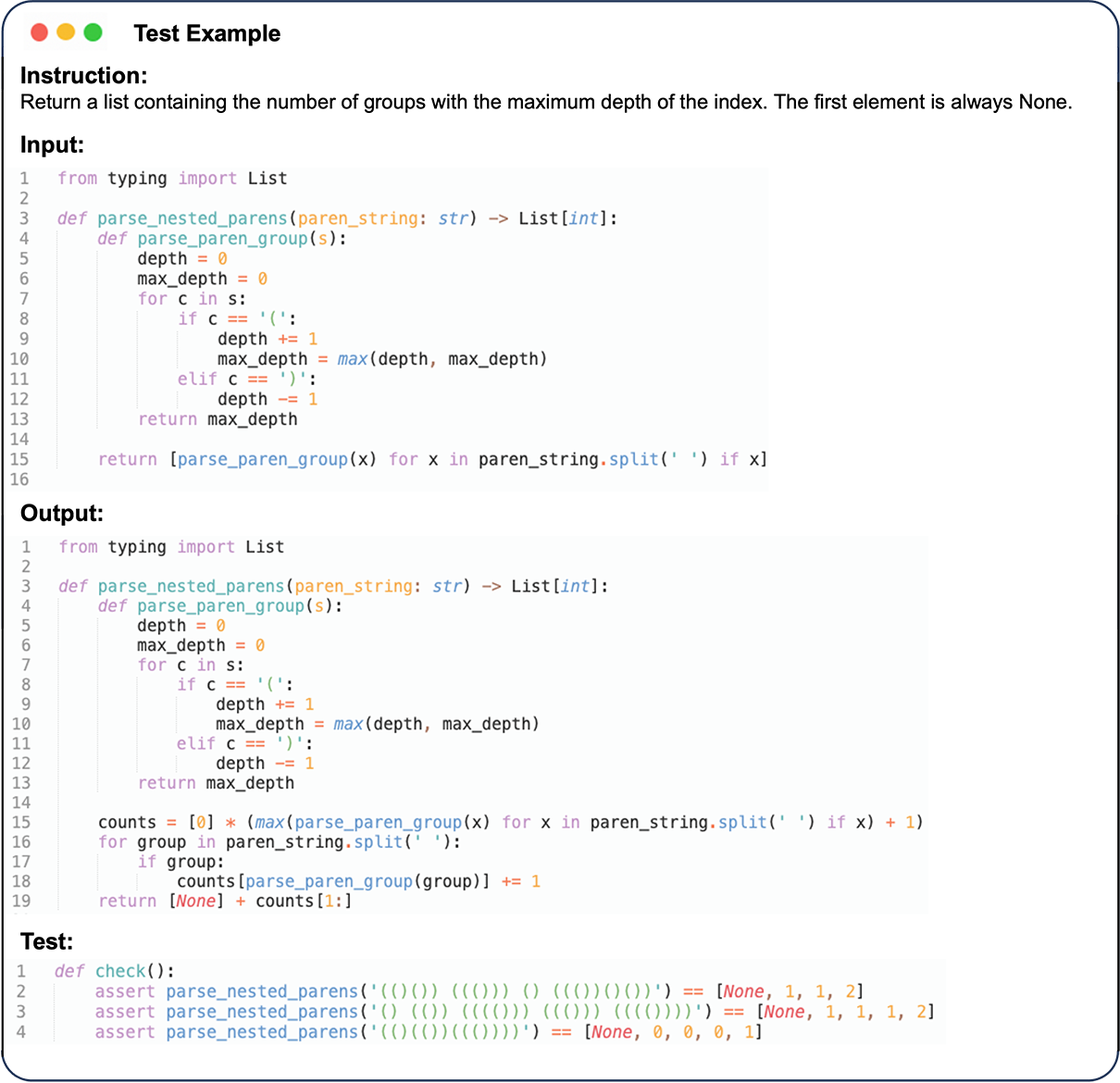}
\caption{An example instance of EditEval.}
\label{fig:editeval_example}
\end{figure*}

\newpage
\section{Comparing Machine-Generated Data and Real-World Data}
\label{appendix:raw_data}
\begin{figure*}[h!]
\centering
\includegraphics[width=0.8\textwidth]{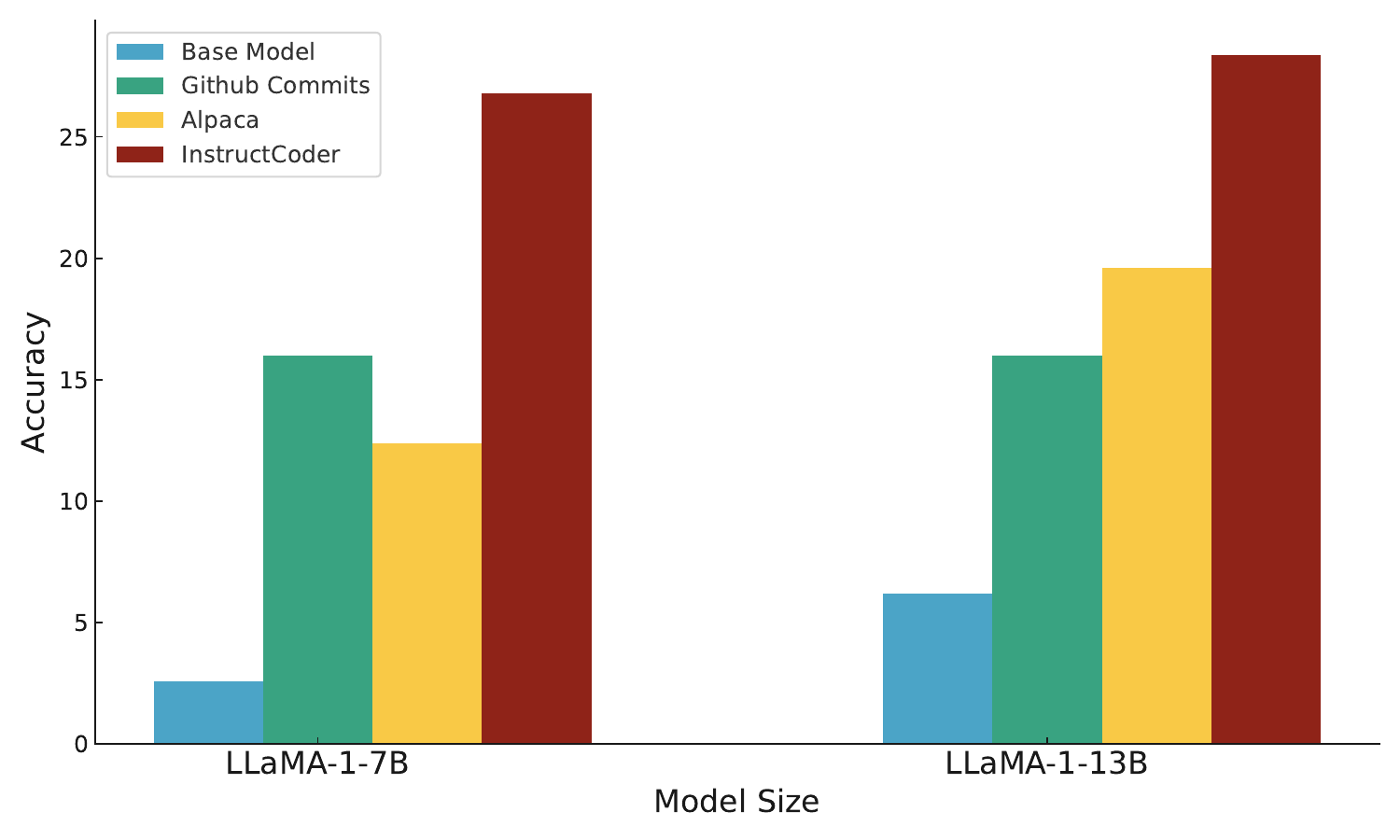}
\caption{EditEval accuracies of instruction fine-tuned LLaMA-1 models (7B and 13B) with GitHub commits and other datasets. InstructCoder significantly outperformed
GitHub commits, and the lead is more pronounced with a larger base model, indicating the effectiveness of
InstructCoder. Conversely, fine-tuning with raw GitHub commits yields poor results, and is the worst among all
three data sources on LLaMA-1 13B.}
\label{tab:raw_github_data}
\end{figure*}

Given the substantial repository of code and commit data available on GitHub, a natural idea is to utilize these real-world data to fine-tune a model to perform code editing. However, as discussed in Section \ref{sec:seed_data_collection}, these data from GitHub can be extremely noisy, especially in the commit messages, rendering them a sub-optimal choice for instruction-tuning. On the other hand, machine-generated data is increasingly recognized for its utility, as evidenced by various studies that achieves enhanced results with this type of data~\citep{gunasekar2023phi1, li2023phi1.5,wang2023improving_embeddings}. This approach provides better controllability over the distribution of the generated contents and facilitates  the collection of diverse data, including those under-represented or difficult to mine and clean from real-world data.

The experiment results in Figure \ref{tab:raw_github_data} corroborate the usage of machine-generated data. We further collected GitHub commits matching the size of InstructCoder, and used the same hyperparameters for instruction fine-tuning. As can be seen in the results, InstructCoder significantly outperformed
raw GitHub commits, and the lead is more profound with a larger base model, demonstrating the effectiveness of
InstructCoder. On the other hand, fine-tuning with GitHub commits yields poor results, and is the worst among
all three data sources on LLaMA-1 13B. The observation suggests that using machine-generated data for instruction fine-tuning is superior in terms of training code editing models.

\newpage
\section{Prompts}
\label{appendix:prompts}
The prompts used in our data collection and experiments are listed in Table \ref{tab:prompts}. 

\begin{table}[h!]
\begin{tabularx}{\textwidth}{|l|X|}
\hline
\textbf{Stage}                  & \textbf{Prompt}                                                                                                                                                                                                                                                                                                                                                                                                                                                             \\ \hline
Instruction Generation & \texttt{Given the existing instructions, please generate a list of diverse Python code editing instructions. The new instructions should address diverse editing tasks. Please ensure that the instructions are clear and diverse. Include any relevant variable names in the instructions.}                                                                                                                                                                                 \\ \hline
Scenario Generation    & \texttt{Given a Python code editing task, please come up with 10 diverse scenarios with concise descriptions of where this task could be performed or come from.}                                                                                                                                                                                                                                                                                                \\ \hline
Instance Generation    & \texttt{Given Python code editing task instructions and their scenarios where the task instruction could be used, you need to come up with examples for the following code editing tasks. You need to generate an input and output code pair and make sure your variable names are suitable for the scenario. The input code is related to the task instruction, but must NOT meet the task requirements. The output code fulfills the task requirements based on the input code.} \\ \hline
GPT4 Evaluation        & \texttt{Given a code editing instruction, please determine if the output is an acceptable edited code response to the instruction and input. Give "Yes" or "No".}   \\
\hline
\end{tabularx}

\caption{Prompts used in this work.}
\label{tab:prompts}
\end{table}

\newpage

\section{Qualitative Examples of Scenario-Conditional Generation }
Three comparisons are presented, each showing instances that were generated with or without the inclusion of a scenario. 
\label{appendix:example-scenario-conditional-generation}
\begin{figure*}[h!]
\centering
\includegraphics[width=\textwidth]{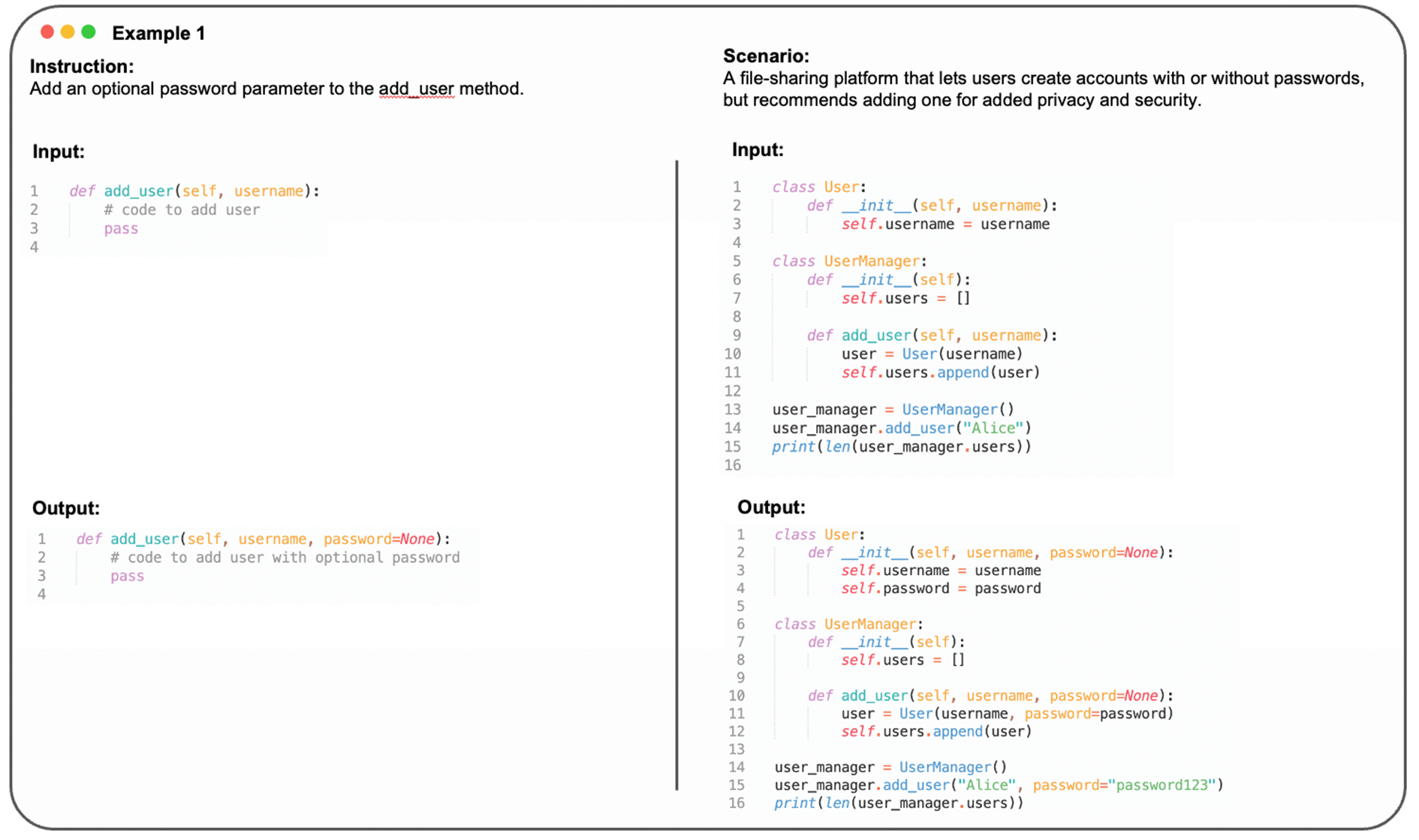}
\caption{Example instance \#1 generated without scenario (Left) and with scenario (Right)}
\label{fig:scenario_example1}
\end{figure*}

\begin{figure*}[h!]
\centering
\includegraphics[width=\textwidth]{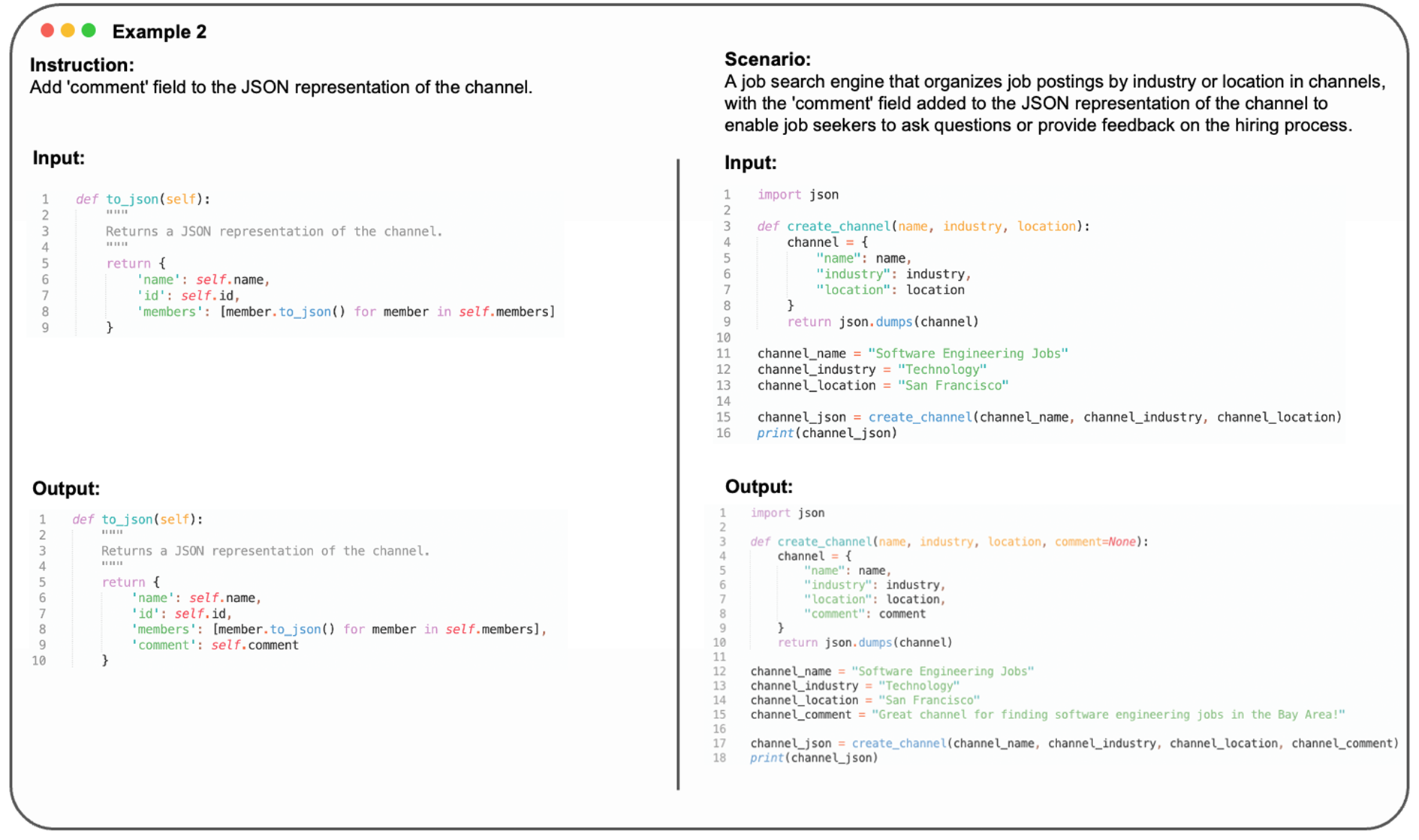}
\caption{Example instance \#2 generated without scenario (Left) and with scenario (Right)}
\label{fig:scenario_example2}
\end{figure*}

\begin{figure*}[h!]
\centering
\includegraphics[width=\textwidth]{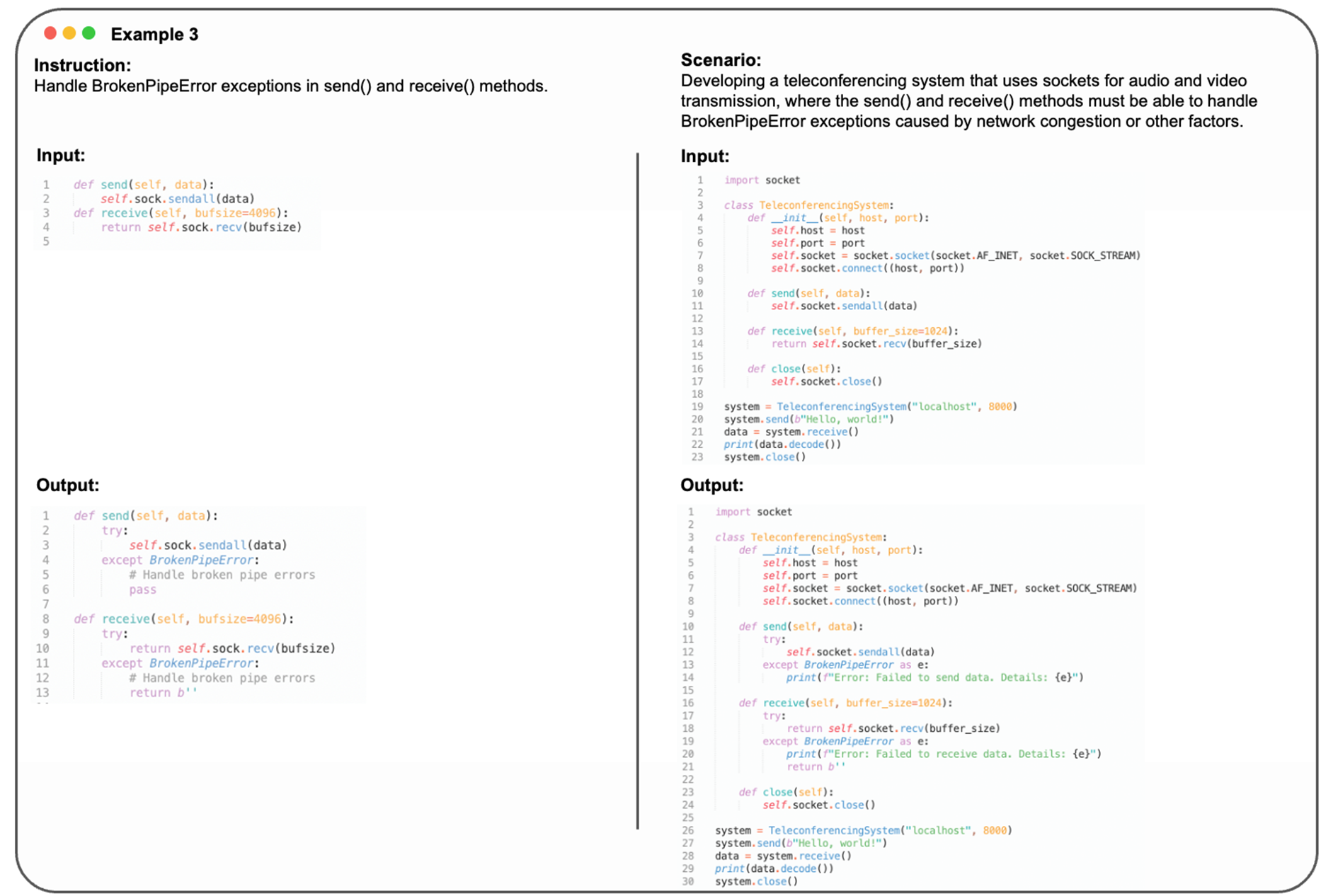}
\caption{Example instance \#3 generated without scenario (Left) and with scenario (Right)}
\label{fig:scenario_example3}
\end{figure*}

\clearpage
\newpage
\section{Additional statistics of InstructCoder}
\label{appendix:additional_stats}

\begin{table*}[h!]
\centering
\begin{tabular}{llll}
\hline
\textbf{Token Length} & \textbf{Instruction} & \textbf{Input} & \textbf{Output}\\
\hline
\verb|mean| & 21.85 & 172.03 & 248.43 \\
\verb|25%| & 17 & 99 & 138 \\
\verb|50%| & 21 & 147 & 213\\
\verb|75%| & 26 & 218 & 321\\
\verb|min| & 3 & 10 & 10\\
\verb|max| & 116 & 1019 & 1024\\
\hline
\end{tabular}
\caption{Token length statistics using the LLaMA~\citep{touvron2023llama} tokenizer.}
\label{tab:statistic}
\end{table*}

\begin{figure*}[h!]
\centering
\includegraphics[width=0.7\textwidth]{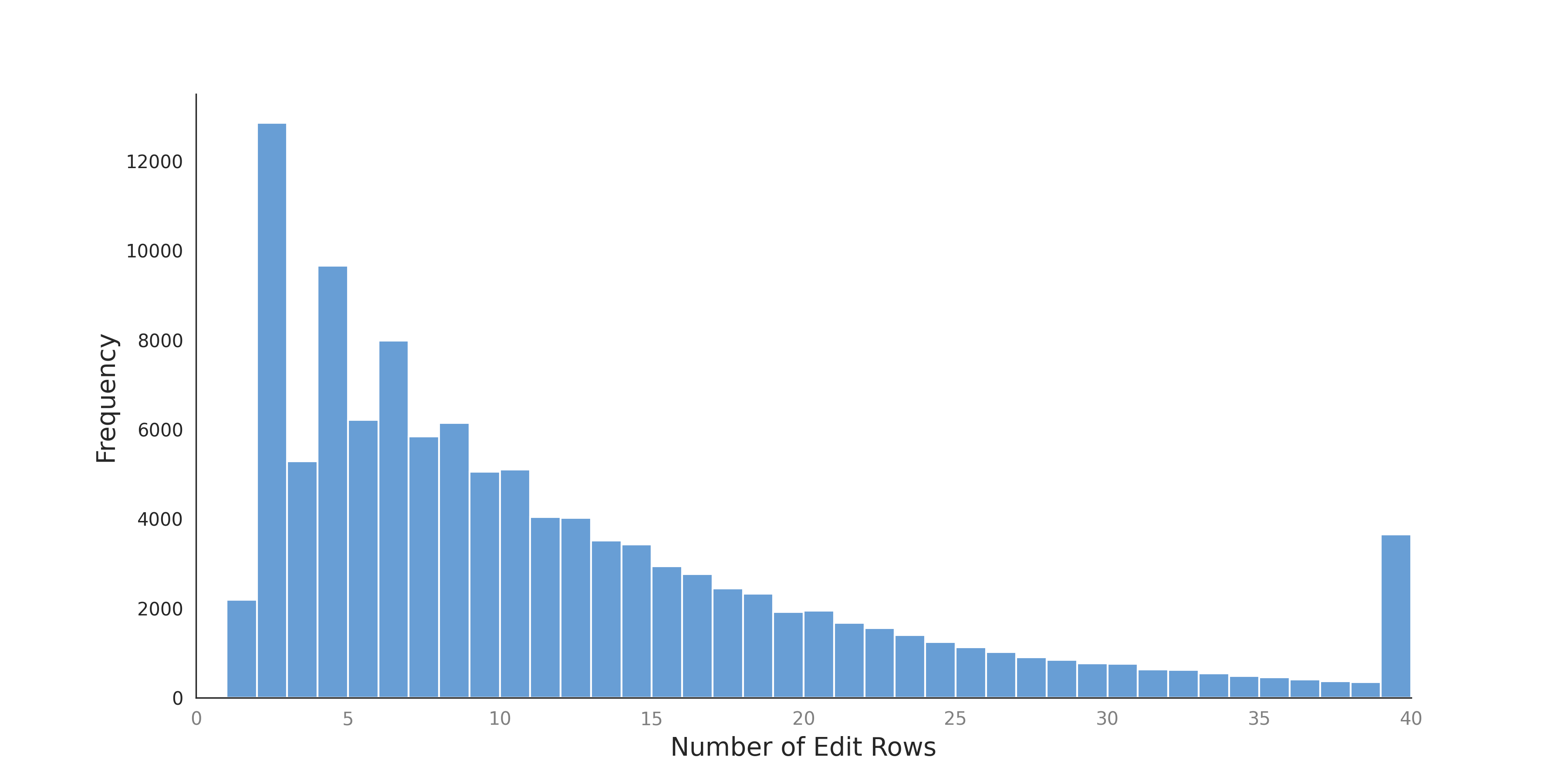}
\caption{Edit rows distribution of InstructCoder. Numbers greater than 40 are aggregated as the last bin.}
\label{fig:edit_distribution}
\end{figure*}


\newpage
\section{Hyperparameters}
\label{appendix:hyperparameters}

The hyperparameters used in all finetuning experiments are listed in Table \ref{tab:hyperparameters}. For all inferences, we utilize greedy decoding. For OpenAI's GPTs, we achieve this by setting its temperature to 0.

\begin{table*}[h!]
\centering
\begin{tabular}{cc}
\toprule
Hyperparameter      & Value                     \\ \midrule
learning rate       & 0.0003                    \\
batch size          & 128                       \\
epochs              & 3                         \\
max sentence length & 1024                      \\
lora rank           & 16                        \\
lora dropout        & 0.05                      \\
lora modules        & key, query, value, output \\ \bottomrule
\end{tabular}

\caption{Hyperparameters used for finetuning language models.}
\label{tab:hyperparameters}
\end{table*}

\section{Qualitative Examples Generated by Finetuned LLaMA-33B}
\label{appendix:qualitative_examples}
We demonstrate some qualitative example responses generated by finetuned LLaMA-33B.

\begin{figure*}[hbt!]
\centering
\includegraphics[width=\textwidth]{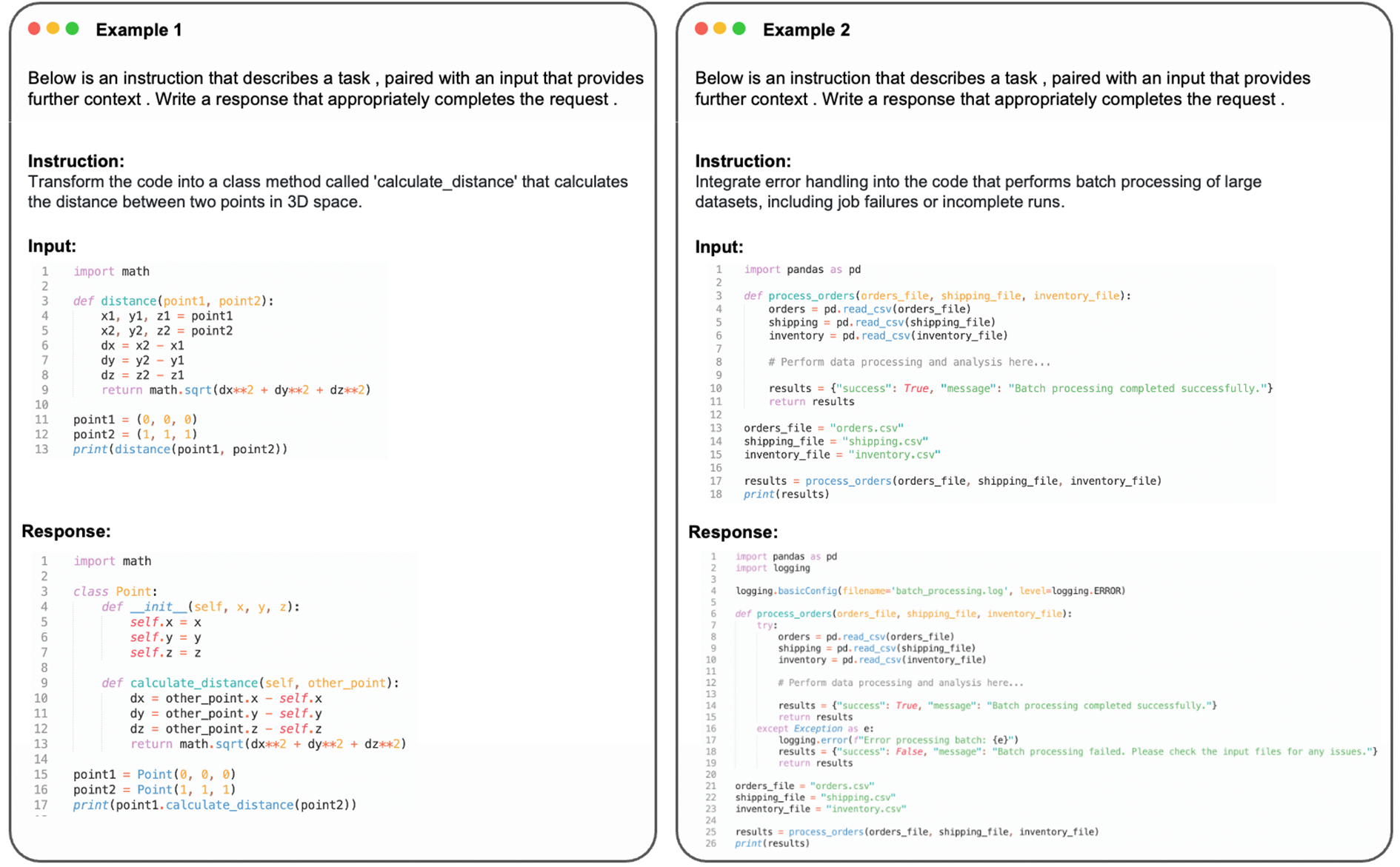}
\caption{Qualitative examples generated by finetuned LLaMA-33B}
\label{fig:qualitative_examples}
\end{figure*}

\newpage
\section{Alignment of GPT-4 Evaluation and Human Evaluation}
\label{appendix:discussion_gpt-4-eval}
Due to the extremely demanding nature of creating automated tests, we seek to investigate the viability of using GPT-4 as an automatic evaluator to lessen the extensive human effort involved. Using LLMs as generation evaluators has been demonstrated effective in NLG tasks~\citep{liu2023gpteval,wang2023chatgpt,fu2023gptscore}, and especially in code generation~\citep{zhuo2023large}. 
To further validate this idea, we collected an additional 134 commits data testing purposes and processed them in the same manner as the seed tasks. Both GPT-4 evaluation and human evaluation are conducted on this dataset to assess their alignment.  

\begin{figure}
    \centering
    \includegraphics[width=0.95\textwidth]{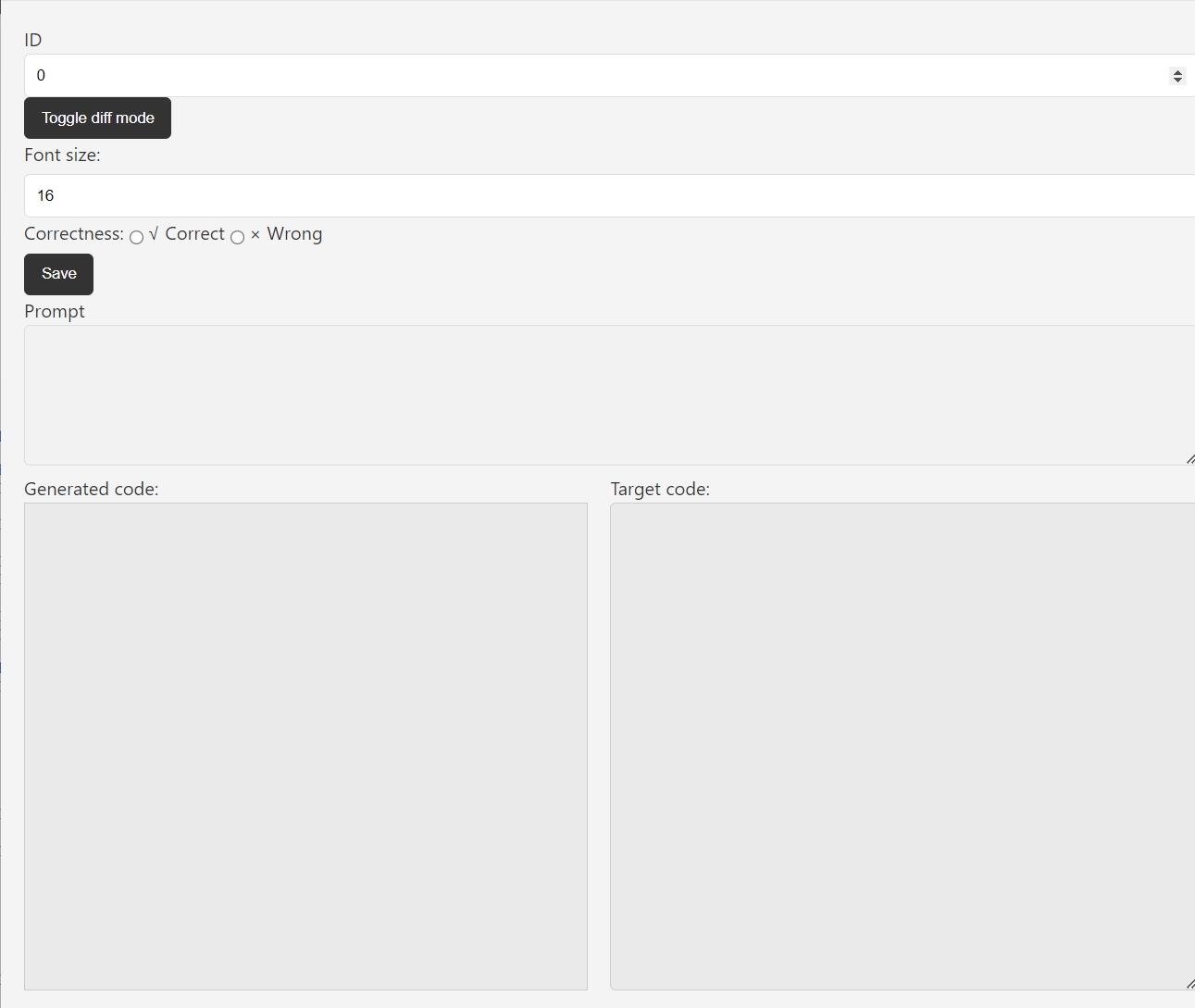}
    \caption{A screenshot of our human scoring annotation tool.}
    \label{fig:anno_tool}
\end{figure}

\paragraph{Human evaluation.} Each sample is annotated by three examiners, and the average accuracy is recorded. We developed an annotation tool to ensure the impartiality of evaluation (see Figure \ref{fig:anno_tool} for the user interface). Generations of different models are shuffled and the anonymity of the models is guaranteed. The edit is annotated as \emph{correct} if it correctly reflects the instruction demands and \emph{wrong} if it fails to follow the instruction. 

\paragraph{GPT-4 evaluation.} We ask GPT-4 to evaluate if the code edit is an acceptable response to the input and collect the correct rate. The prompts for GPT-4 evaluation can be found in \ref{appendix:prompts}.

\paragraph{Results.} We carry out the experiments on the code edits generated by ChatGPT and LLaMA of three sizes fine-tuned with InstructCoder. While we found that the human annotators are always slightly stricter than the GPT-4 evaluator, the overall Cohen's Kappa value of the GPT-4 evaluations and human evaluations reaches 0.665, which is substantial according to ~\citet{cohen1960cohenkappa}. This renders GPT-4 evaluation as a convenient and effective method for evaluating the correctness of code edit tasks.

\newpage

\section{Data Filtering Process}
The detailed process of filtering the dataset is listed below:
\begin{itemize}
    \item We selected GitHub repos with over 100 stars to ensure the overall quality. We only utilized repos with permissive licenses (MIT, Apache-2.0, GPL-3.0, GPL-2.0, BSD-2.0, BSD-3.0, LGPL-2.1, LGPL-3.0, AGPL-3.0).
    \item We kept commits in which only one single .py file was changed.
Using git-diff, we identified and preserved commits where only one code block was changed.
    \item We discarded commits with single-word or empty commit messages.
    \item We removed commits with over 100 edited rows.
\end{itemize}
Manual:

\begin{itemize}
    \item We discarded rare commits containing inappropriate language.
    \item We discarded commits where the change in the source code does not match the commit message.
    \item We filtered out project-specific adjustments that lack sufficient context.
    \item We utilized Codex~\citep{chen2021codex} to rewrite ambiguous commit messages, enhancing the clarity of the intended code edits.
\end{itemize}

\end{document}